\documentclass[10pt,conference]{IEEEtran}
\usepackage{amsmath}
\usepackage{cite}
\usepackage{graphicx}
\usepackage{amsfonts}
\usepackage{latexsym}
\usepackage{subfigure}
\usepackage{algorithm}
\usepackage{algorithmic}
\usepackage{color}
\usepackage{times}
\usepackage{epsfig, graphics}
\usepackage{bm}
\usepackage{subfigure}
\usepackage{graphicx,wrapfig,lipsum}
\usepackage{booktabs}
\begin{document}

\newtheorem{theorem}{Theorem}
\newtheorem{lemma}{Lemma}
\newtheorem{conjecture}{Conjecture}
\newtheorem{corollary}{Corollary}
\newtheorem{definition}{Definition}
\newtheorem{scheme}{Scheme}
\newcommand{\argmax}{\arg\!\max}
\newcommand{\pound}{\operatornamewithlimits{\gtrless}}
\IEEEoverridecommandlockouts

\newenvironment{changemargin}[2]{%
\begin{list}{}{%
\setlength{\topsep}{0pt}%
\setlength{\leftmargin}{#1}%
\setlength{\rightmargin}{#2}%
\setlength{\listparindent}{\parindent}%
\setlength{\itemindent}{\parindent}%
\setlength{\parsep}{\parskip}%
}%
\item[]}{\end{list}}

\title{Generative Adversarial Networks for Black-Box API Attacks with Limited Training Data} 

\author{\IEEEauthorblockN{Yi Shi, Yalin E. Sagduyu, Kemal Davaslioglu, and Jason H. Li}

\IEEEauthorblockA{\\Intelligent Automation, Inc. \\ Rockville, MD 20855, USA\\
Email:\{yshi, ysagduyu, kdavaslioglu, jli\}@i-a-i.com}
}

\maketitle

\begin{changemargin}{0.075in}{0.075in}

\begin{abstract}
As online systems based on machine learning are offered to public or paid subscribers via application programming interfaces (APIs), they become vulnerable to frequent exploits and attacks. This paper studies adversarial machine learning in the practical case when there are rate limitations on API calls. The adversary launches an exploratory (inference) attack by querying the API of an online machine learning system (in particular, a classifier) with input data samples, collecting returned labels to build up the training data, and training an adversarial classifier that is functionally equivalent and statistically close to the target classifier.
The exploratory attack with limited training data is shown to fail to reliably infer the target classifier of a real  text classifier API that is available online to the public.
In return, a generative adversarial network (GAN) based on deep learning is built to generate synthetic training data from a limited number of real training data samples, thereby extending the training data and improving the performance of the inferred classifier. 
The exploratory attack provides the basis to launch the causative attack (that aims to poison the training process) and evasion attack (that aims to fool the classifier into making wrong decisions) by selecting training and test data samples, respectively, based on the confidence scores obtained from the inferred classifier. These stealth attacks with small footprint (using a small number of API calls) make adversarial machine learning practical under the realistic case with  limited training data available to the adversary. 
\end{abstract}
\section{Introduction}
\emph{Artificial intelligence} (AI) finds diverse and far-reaching applications ranging from cyber security and intelligence analysis to Internet of Things (IoT) and autonomous driving. To support AI, \emph{machine learning} provides computational resources with the ability to learn without being explicitly programmed. Enabled by hardware accelerations for computing, the next generation machine learning systems such as those based on \emph{deep learning} are effectively used for different classification, detection, estimation, tracking, and prediction tasks.

Machine learning applications often involve sensitive and proprietary information such as training data, machine learning algorithm and its hyperparameters, and functionality of underlying tasks. As such applications become more ubiquitous, machine learning itself becomes subject to various exploits and attacks that expose the underlying tasks to unprecedented threats. \emph{Adversarial machine learning} is an emerging field that studies machine learning in the presence of an \emph{adversary} \cite{Papernot16}. In one prominent example adversarial images were generated by slightly perturbing images to fool a state-of-the-art image classifier, e.g., a perturbed panda image was recognized as a gibbon by the classifier \cite{Goodfellow14}. 

Machine learning systems are typically offered to public or paid subscribers via an application programming interface (API), e.g., Google Cloud Vision \cite{Google:Vision}, 
provide cloud-based machine learning tools to build and train machine learning models. This online service paradigm makes machine learning vulnerable to \emph{exploratory (or inference) attacks} \cite{Barreno06, Tramer16, Papernot17, Shi17:DL,  Wu16, bookchapter2018, Yi2018} that aim to understand how the underlying machine learning algorithm works. 
An exploratory attack can be launched as a black-box attack \cite{Tramer16, Papernot17, Shi17:DL} without any prior knowledge on the machine learning algorithm and the training data. In this attack, the adversary  calls the target classifier $T$ with a large number of samples, collects the labels, and then uses this data to train a deep learning classifier $\hat T$ that is functionally equivalent (i.e., statistically similar) to the target classifier $T$. This attack infers the full functionality of the classifier and implicitly steals its underlying training data,  machine learning algorithm, and hyperparameter selection.

After an adversary infers $\hat T$, it can launch further attacks such as \emph{causative attacks} \cite{Pi16, Alfed16, Shi17:Milcom} and \emph{evasion attacks} \cite{Shi17:Milcom, Papernot16:evasion,  Kurakin16, Moosavi-Dezfooli15:DeepFool}.
In a causative attack, the adversary provides the target classifier with incorrect information as additional training data to reduce the reliability of the retrained classifier. In particular, the adversary runs some samples through the inferred classifier $\hat T$. If their deep learning scores (likelihood of labels) are far away from the decision boundary, the adversary changes their labels and sends these mislabeled samples as training data to the target classifier $T$.
In an evasion attack, the adversary provides the target classifier with test data that will result in incorrect labels (e.g.,  the classifier is fooled into accepting an adversary as legitimate). In particular, the adversary runs some samples through the inferred classifier $\hat T$. If their deep learning scores are confined within the decision region for label $j$ and are close to the decision region for the other label $i$, the adversary feeds those samples to the target classifier $T$ that is more likely to classify their labels to $j$ instead of $i$.

One major limitation of exploratory attacks that has been overlooked so far is the extent to which training data can be collected from the target application.  Usually, APIs have \emph{strict rate limitations} on how many application samples can be collected over a time period (ranging from a limited number samples per second to per day). We consider the realistic case that the number of calls is limited due to various reasons, e.g., the target classifier $T$ may limit the number of calls that a user can make, or identify a large number of calls as malicious by a defense mechanism.

In this paper, we select a real online text analysis API, DatumBox \cite{Datumbox}, as the target classifier $T$. The underlying machine learning algorithm is unknown to the adversary that treats $T$ as a black box.  First, we show that as the API call rate is limited, the adversary cannot reliably infer $T$ using the exploratory attack. Second, we present an approach to enhance the training process of the exploratory attack.
This approach is based on the \emph{generative adversarial network} (GAN) \cite{Goodfellow14:GAN} to augment training data with synthetic samples.
We show that the exploratory attack with the GAN is successful even with a small number of training samples (acquired from the API of the target classifier $T$) and the inferred classifier $\hat T$ based on the deep neural network is statistically close to $T$.
The GAN consists of a generator $G$ and a discriminator $D$ playing a minimax game. $G$ aims to generate realistic data (with labels), while $D$ aims to distinguish data generated by $G$ as real or synthetic.
By playing a game between $G$ and $D$, the GAN generates additional training data without further calls to $T$. With this additional training data, $\hat T$ is better trained.
We show that the proposed approach significantly improves the performance of $\hat T$, which is measured as the statistical difference of classification results, i.e., labels returned, by $T$ and $\hat T$.
\begin{figure}
\centering
\includegraphics[width=0.85\columnwidth]{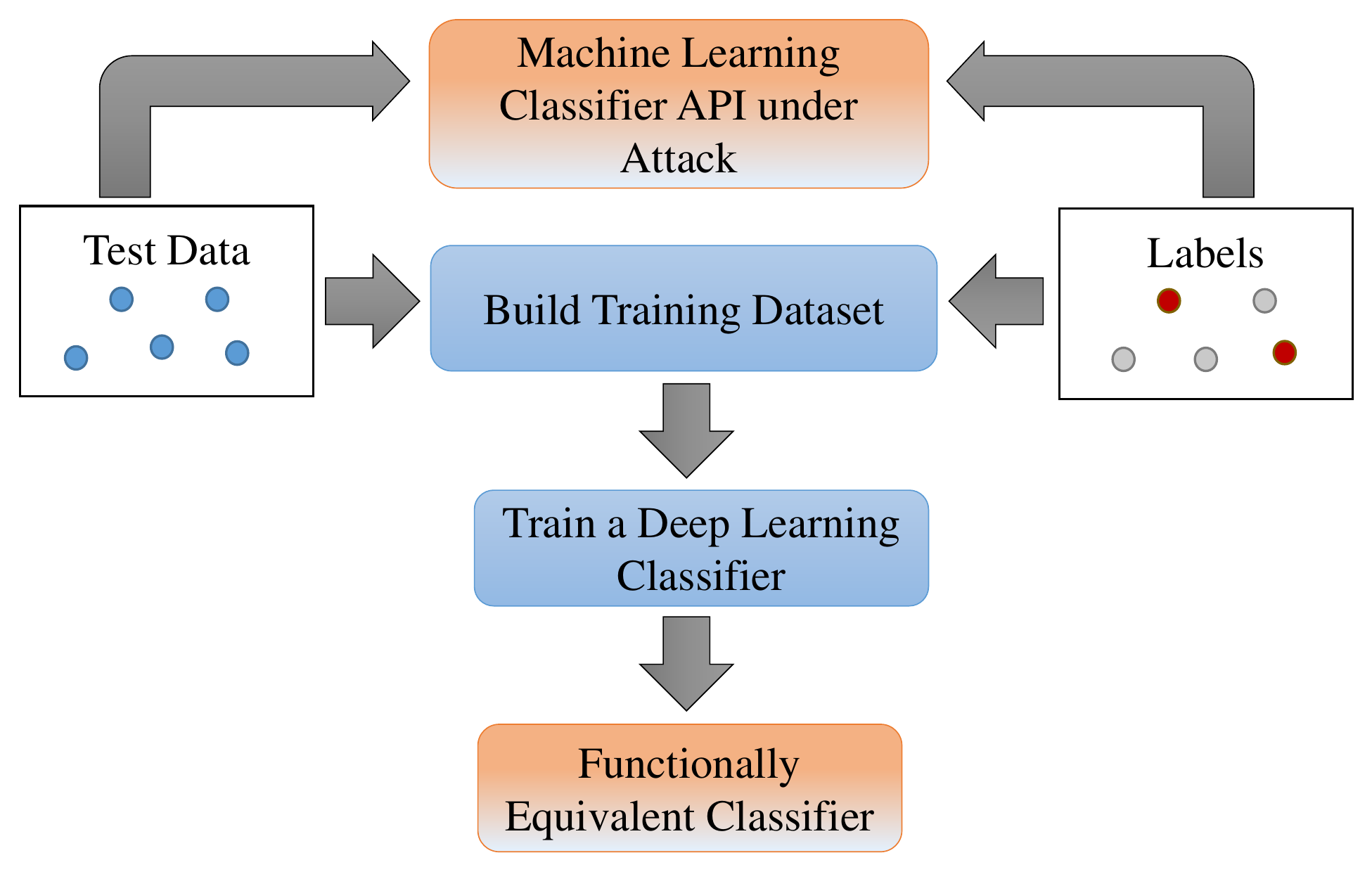}
\vspace{-0.1cm}
\caption{Adversarial deep learning with limited training data.}\label{fig:systemfigure}\
\vspace{-0.5cm}
\end{figure}

Adversarial machine learning with limited training data was studied in \cite{Papernot17} from a different perspective, where the adversary first infers the target classifier with limited real training data and then generates synthetic data samples in an evasion attack by adding adversarial perturbations to real data that are gradually improved with labels queried from the target classifier.
Our setting is different in several aspects. First, the size of training data used in \cite{Papernot17} (6,400 samples) is much larger than ours (100 samples). Second, we select real samples that the target will misclassify, instead of altering real samples. Third, we consider a text classifier as the target classifier, whereas \cite{Papernot17} considered an image classifier. Fourth, we improve the inferred classifier (not the adversarial samples) that is used later to provide an arbitrary number of selected samples to evasion or causative attacks without further querying the target classifier.

Data augmentation with the GAN was studied for the low-data regime, where synthetic data samples were generated for image applications in \cite{DAGAN} and wireless communications in \cite{Kemal2018} and \cite{Tugba2018}. In this paper, we used the GAN to support attacks by augmenting the training data for an adversary that operates with strict rate limitations on the API calls made to a target text classifier. A similar setting was considered in \cite{Yihst2018}. Instead of adding synthetic data samples, \cite{Yihst2018} used active learning to reduce the number of API calls needed to obtain the labels of real data samples.
The rest of the paper is organized as follows.
Section~\ref{sec:Datumbox} studies the exploratory attack on an online classifier with a large number of API calls. Training data augmentation with the GAN is presented in Section \ref{sec:gan} to launch the exploratory attack with a limited number of API calls.
 The subsequent causative and evasion attacks are studied in Section \ref{sec:ceattacks}.
Section~\ref{sec:conclusion} concludes the paper.

\section{Black-Box Exploratory Attack}
\label{sec:Datumbox}

The system model is shown in Fig.~\ref{fig:systemfigure}. There are various online APIs for machine learning services with either free subscriptions or paid licenses. For example, DatumBox provides a number of machine learning APIs for text analysis \cite{Datumbox}. A user can write its own code to call these APIs, specify some input text data, and obtain the returned labels, e.g., subjective or objective. These services are often based on sophisticated algorithms (e.g., machine learning) and tuned by a large amount of training data, which requires a significant effort. 
The number of calls made by a user depends on the license and is limited in general, e.g., 1000 calls per day for free license of DatumBox. The adversary launches a black-box exploratory attack. The target classifier $T$ is based on an algorithm (e.g., Naive Bayes or Support Vector Machine (SVM) or a more sophisticated neural network) that is unknown to the adversary. In addition, the adversary does not know the training data that has been used to train the target classifier $T$. The adversary pursues the following steps:
\begin{enumerate}
\item The adversary sends a set $S$ of samples to the target classifier $T$ and receives the label $T(s)$ returned by $T$ for each sample $s \in S$.
\item The adversary trains a deep neural network using the training data from Step 1 and builds its own inferred classifier $\hat T$.
\end{enumerate}

The adversary applies deep learning (based on a multi-layered neural network) to infer $T$. Deep learning refers to training a deep neural network for computing some function. Neural network consists of \emph{neurons} that are joined by weighted connections (\emph{synapses}). In a \emph{feedforward neural network} (FNN) architecture, neurons are arranged in layers. A signal $x_k$ at the input of synapse $k$ is connected to neuron $j$ and multiplied by the synaptic weight $w_{jk}$. An adder sums these weighted multiplications and inputs the sum to an activation function denoted by $\sigma(\cdot)$. The bias $b_j$ of the $j$th layer increases or decreases the sum that is input of the activation function. These operations can be expressed as ${y}_j = \sigma\left(\sum_{k=1}^m {w}_{jk} {x}_{k} + b_j\right)$. The operation of an FNN is illustrated in Fig.~\ref{fig:synapses} for one of the layers
with
$m$ neurons. 
The weights and biases are determined by the backpropagation algorithm. 

\begin{figure}[t!]
	\centering
	\includegraphics[width=0.6\columnwidth]{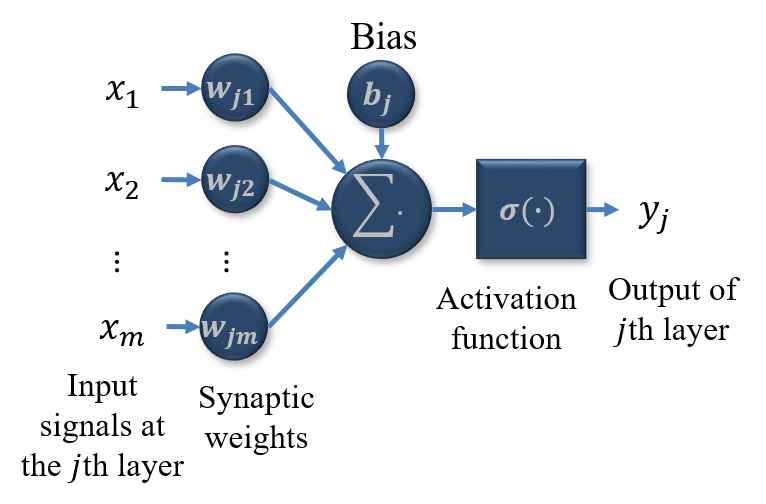}
	\caption{One of the neural network layers.}\label{fig:synapses}
\end{figure}

We consider an exploratory attack on a real online classifier, namely the text subjectivity analysis API of DatumBox \cite{Datumbox}. The adversary builds training data by calling this API with different text samples and collecting the returned labels (subjective or objective). There are two types of labels. Label $1$ includes  subjective text and label $2$ includes objective text. A user is allowed to make 1000 calls per day with a free license, i.e., one can collect 1000 samples with labels per day.  The Twitter API is used to collect random tweets. These tweets are first preprocessed by removing the stop words, web links (e.g., http and https links), and punctuations. Then, word tokens are created from the cleaned tweets. The word count frequencies of top 20 words that the adversary prepares are presented in Fig.~\ref{fig:wordcount}. Each token is represented by a $k$-dimensional feature vector.

\begin{figure}[t!]
	\centering
	\includegraphics[width=0.8\columnwidth]{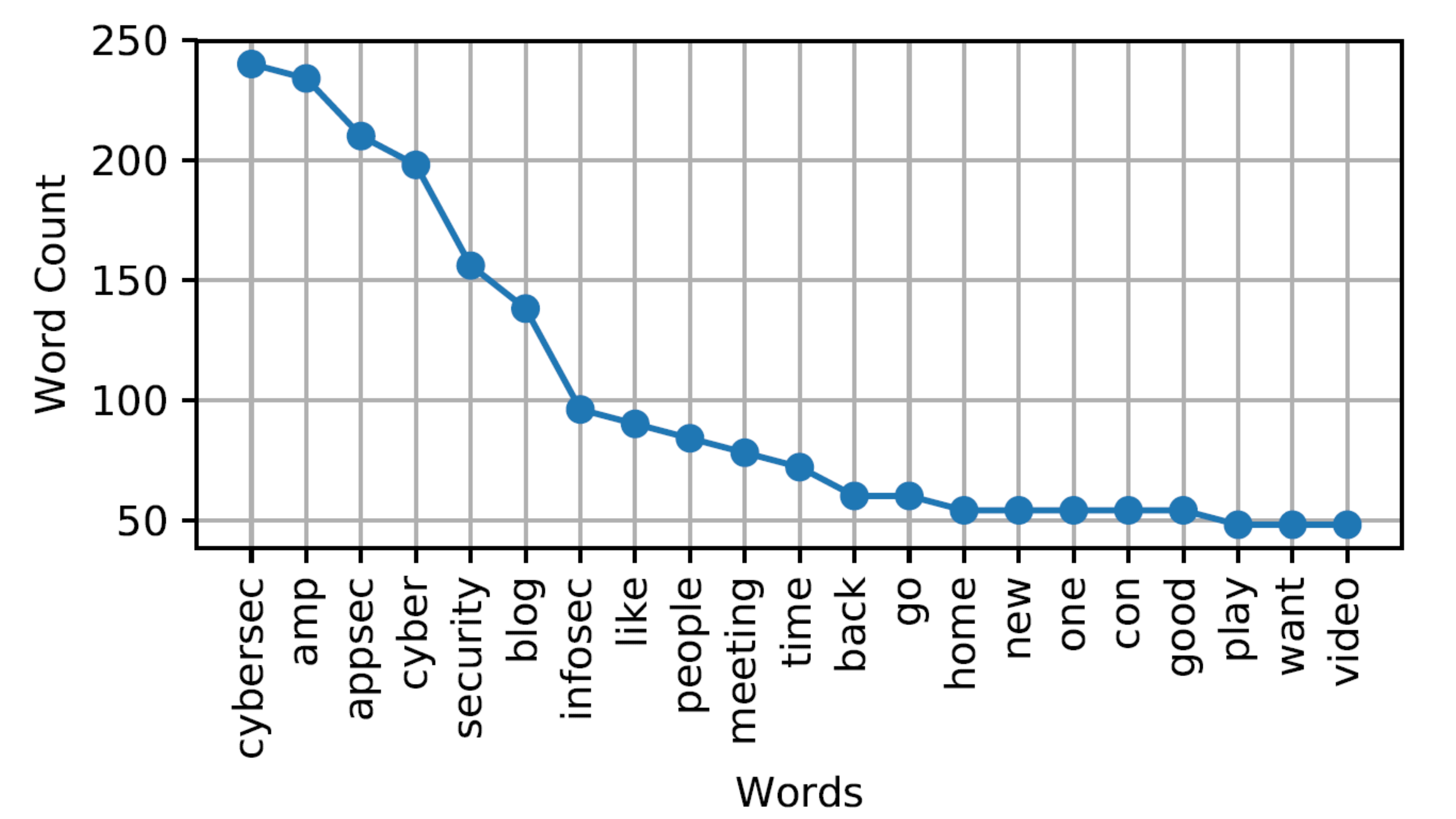}
	\caption{Word count frequencies of top 20 words in text data.}\label{fig:wordcount}
\end{figure}

The inferred classier is built with \emph{deep learning} that requires significantly more data compared to standard machine learning algorithms (e.g., Naive Bayes and SVM). To train a deep neural network,  the weights and biases of the multi-layered neural network are selected. In the meantime, its hyperparameters such as the number of layers and the number of neurons per layer are tuned. To build a set of features as inputs for each text sample, the adversary first obtains a list of words, sorted by their frequencies of occurrence in text such that $W=(w_1, w_2, \cdots)$, where the frequency of occurrence for word $w_i$ is $p_i$ and $p_1 \ge p_2 \ge \cdots$. Then, for each sample of text, the adversary counts the number of occurrence $o_i$ for each word $w_i$ and builds a set of features $(o_1, o_2, \cdots)$. Overall, 1000 features are obtained by considering distributions of top 1000 words. These features are used by the adversary to train the deep learning classifier $\hat T$.

The following two differences between $\hat T$ and $T$ are computed. The difference $d_1(\hat T, T)$ between $\hat T$ and $T$ on label $1$ is the number of samples with ${\hat T}(s)=2$ and $T(s)=1$ divided by the number of samples with label $1$ in test data, and the difference $d_2(\hat T, T)$ between $\hat T$ and $T$ on label $2$ is the number of samples with ${\hat T}(s)=1$ and $T(s)=2$ divided by the number of samples with label $2$ in test data.

The adversary builds the optimal $\hat T$ by selecting hyperparameters (e.g., the number of layers and the number of neurons per layer) to minimize $d_{\max}(\hat T, T) = \max\{d_1(\hat T, T),d_2(\hat T, T)\}$ based on the training data, thereby balancing the false alarm and misdetection errors. We used the Microsoft Cognitive Toolkit (CNTK)  to train the FNN and developed the Python code for hyperparameter optimization. The adversary collects 6000 samples over time and uses half of them as training data and half of them as test data to train a classifier $\hat T$.
We optimize the hyperparameters of the deep neural network as follows:
\begin{itemize}
\item The number of hidden layers is 2.
\item The number of neurons per layer is 30.
\item The loss function is cross entropy.
\item The activation function in hidden layers is sigmoid.
\item The activation function in output layer is softmax.
\item All weights and biases are not initially scaled.
\item Input values are unit normalized in the first training pass.
\item The minibatch size is 20.
\item The momentum coefficient to update the gradient is 0.9.
\item The number of epochs per time slot is 10.
\end{itemize}
The difference between labels returned by $T$ and $\hat T$ is found as $d_1(\hat T, T) = 25.66\%$, $d_2(\hat T, T) = 26.04\%$, and $d(\hat T, T) = 25.80\%$. The difference can be further reduced with more data collected over additional days.  

In this attack, the adversary needs to collect training data by calling the target classifier many times over multiple days (with 1000 samples per day). First, such an attack incurs significant delay. Second, a simple mechanism that limits the number of allowed calls would force this attack to take over a long time.
Such a long-term attack can be detected over time and identified as malicious behavior, and the adversary can be blocked. \emph{Therefore, it is essential that the adversary performs the exploratory attack with a limited number of API calls.} In the next  section, we will present a novel approach to enable a successful exploratory attack with a limited number of training data samples.

\section{Training Data Augmentation with the GAN}
\label{sec:gan}

\begin{figure}
	\centering
	\includegraphics[width=0.85\columnwidth]{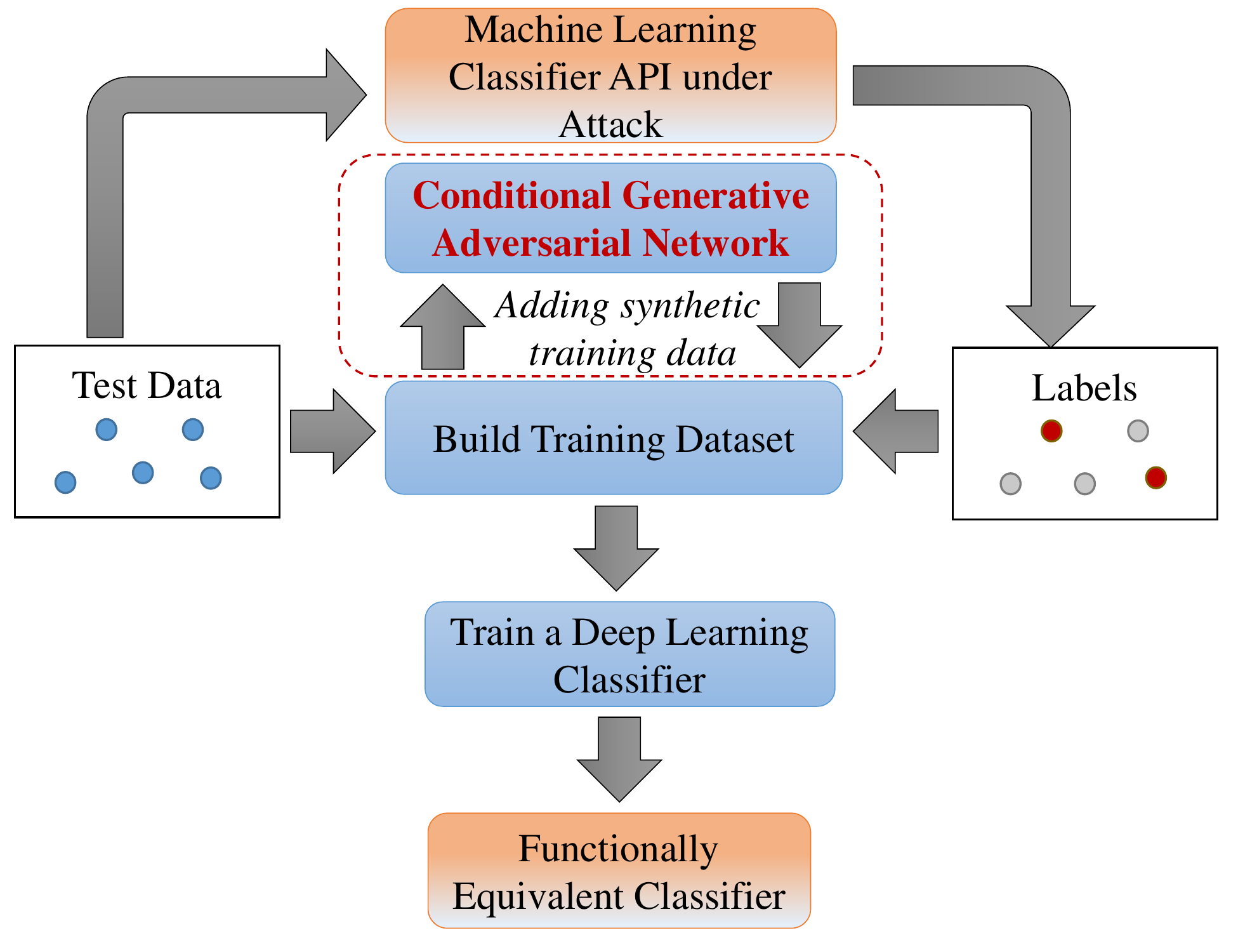}
	\caption{Adversarial deep learning with GAN-based data augmentation.}\label{fig:systemfiguregan}
\end{figure}

Consider the adversarial deep learning system model in Fig.~\ref{fig:systemfiguregan}, which extends the exploratory attack in Section~\ref{sec:Datumbox} by utilizing a GAN to generate synthetic data samples and augment the training data set to train the classifier better without additional calls.
Data augmentation can be performed by using neural networks, e.g.,
generating a new sample by injecting noise to the input of a neural network and dropout layers without changing the labels. Recently, the GAN \cite{Goodfellow14:GAN} was proposed to generate synthetic data and was shown to outperform prior methods due to the better quality of generated synthetic  data that cannot be reliably distinguished from real data \cite{Goodfellow14:GAN, Mirza14:GAN, Arjovsky17}. An example of the GAN is illustrated in Fig.~\ref{fig:gan}.
In a GAN, there are two neural networks, namely the {\em generator\/} and the {\em discriminator\/}, playing a minimax game:
\begin{eqnarray}
 \min_{G} \max_{D} \hspace{0.5em} && \mathbb{E}_{\bm{x} \sim p_{data}} [\log(D(\bm{x}))] \nonumber \\
 && -\mathbb{E}_{\bm{z} \sim p_{\bm{z}}} [\log(1 - D(G(z)))],
\label{eqn:gan_obj}
\end{eqnarray}
where $\bm{z}$ is a noise input to generator $G$ with a model distribution of $p_{\bm{z}}$ and $G(\bm{z})$ is the generator output. Input data $\bm{x}$ has distribution $p_{data}$ and discriminator $D$ distinguishes between the real and generated samples.
Both $G$ and $D$ are trained with backpropagation of error. However, when the generator is trained with the objective in (\ref{eqn:gan_obj}), the gradients of the generator rapidly vanish that makes the training of the GAN very difficult. To address the vanishing gradient problem,  the following objective function at the generator \cite{Arjovsky17} is used:
\begin{align}
\max_G \hspace{.5em} \mathbb{E}_{\bm{z} \sim p_{\bm{z}}} [\log(D(G(\bm{z})))].
\end{align}
Thus, in the first step, we have the discriminator $D$ trained to distinguish between the real and synthetic data.
Then generator $G$
takes random noise as input and maximizes the probability of $D$ making a mistake and fools $D$ by creating samples that resemble the real data in the second step. The conditional GAN extends the GAN concept such that the generator can generate synthetic data samples with labels \cite{Mirza14:GAN}.
The objective function in the conditional GAN is similar to the one of a regular GAN, but as illustrated in Fig.~\ref{fig:cgan}, the terms $D(\bm{x})$ and $D(G(\bm{z}))$ are replaced by $D(\bm{x},\bm{y})$ and $D(G(\bm{z},\bm{y}))$, respectively, to accommodate the labels $\bm{y}$ as conditions. We assume that the adversary uses the conditional GAN.

\begin{figure}[t!]
	\centering
	\includegraphics[width=0.65\columnwidth]{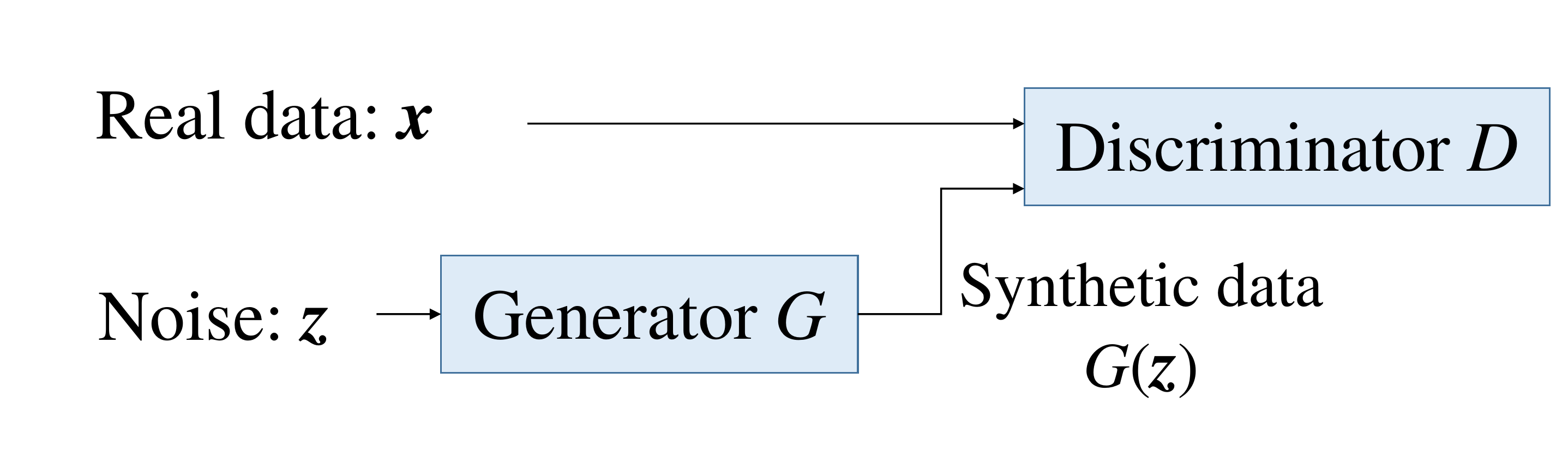}
	\caption{GAN for training data augmentation.}\label{fig:gan}
\end{figure}

\begin{figure}[t!]
	\centering
	\includegraphics[width=0.65\columnwidth]{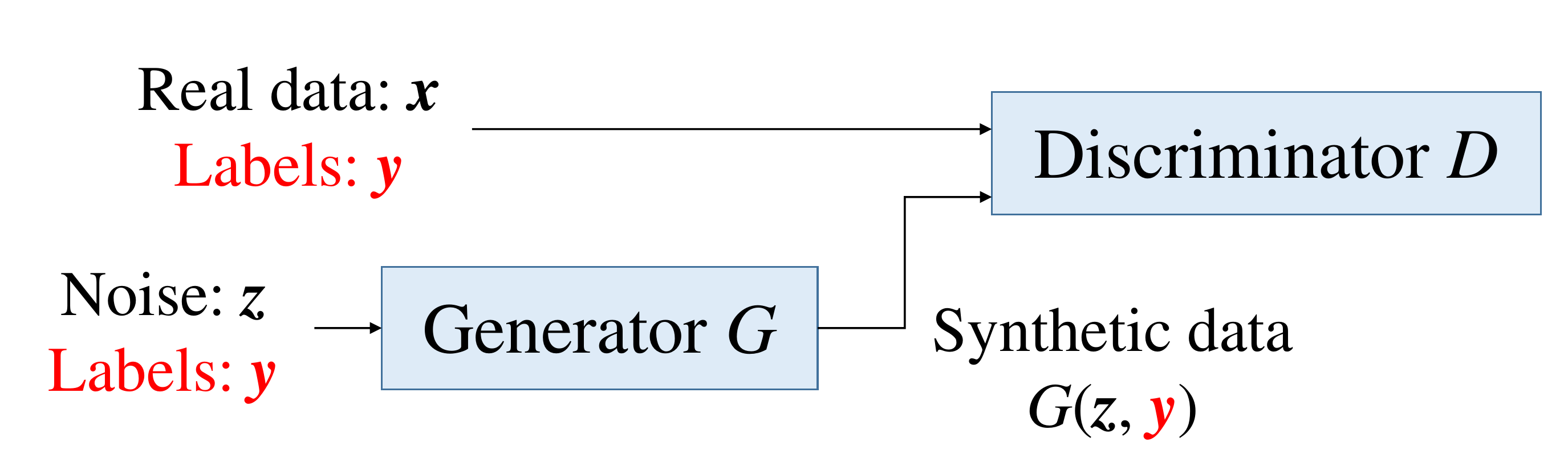}
	\caption{Conditional GAN for training data augmentation.}\label{fig:cgan}
\end{figure}

\begin{table*}
	\caption{Performance (difference between labels returned by $T$ and $\hat T$) with real and synthetic training data.}
	\centering
	\small
	\begin{tabular}{c|c|c|c|c|c}
		\toprule
		Training data size & Number of & Number of & \multicolumn{3}{|c}{Performance}  \\
		(total number of samples) & real samples ($N_r$) & synthetic samples ($N_s$) & $d_1(\hat T, T)$ & $d_2(\hat T, T)$ & $d(\hat T, T)$ \\ \hline
		$100$ & $100$ & $0$ & $44.59\%$ & $45.64\%$ & $45.00\%$ \\ \hline
		$150$ & $100$ & $50$ & $32.79\%$ & $28.72\%$ & $31.20\%$ \\ \hline
		$200$ & $100$ & $100$ & $27.21\%$ & $27.69\%$ & $27.40\%$ \\ \hline
		$250$ & $100$ & $150$ & $30.49\%$ & $26.67\%$ & $29.00\%$ \\ \hline
		$300$ & $100$ & $200$ & $28.52\%$ & $28.21\%$ & $28.40\%$ \\ \hline
		$400$ & $100$ & $300$ & $31.48\%$ & $32.31\%$ & $31.80\%$ \\
		\bottomrule
	\end{tabular}
	\label{tab:gan}
\end{table*}

\begin{figure}[t!]
	\centering
	\includegraphics[width=1\columnwidth]{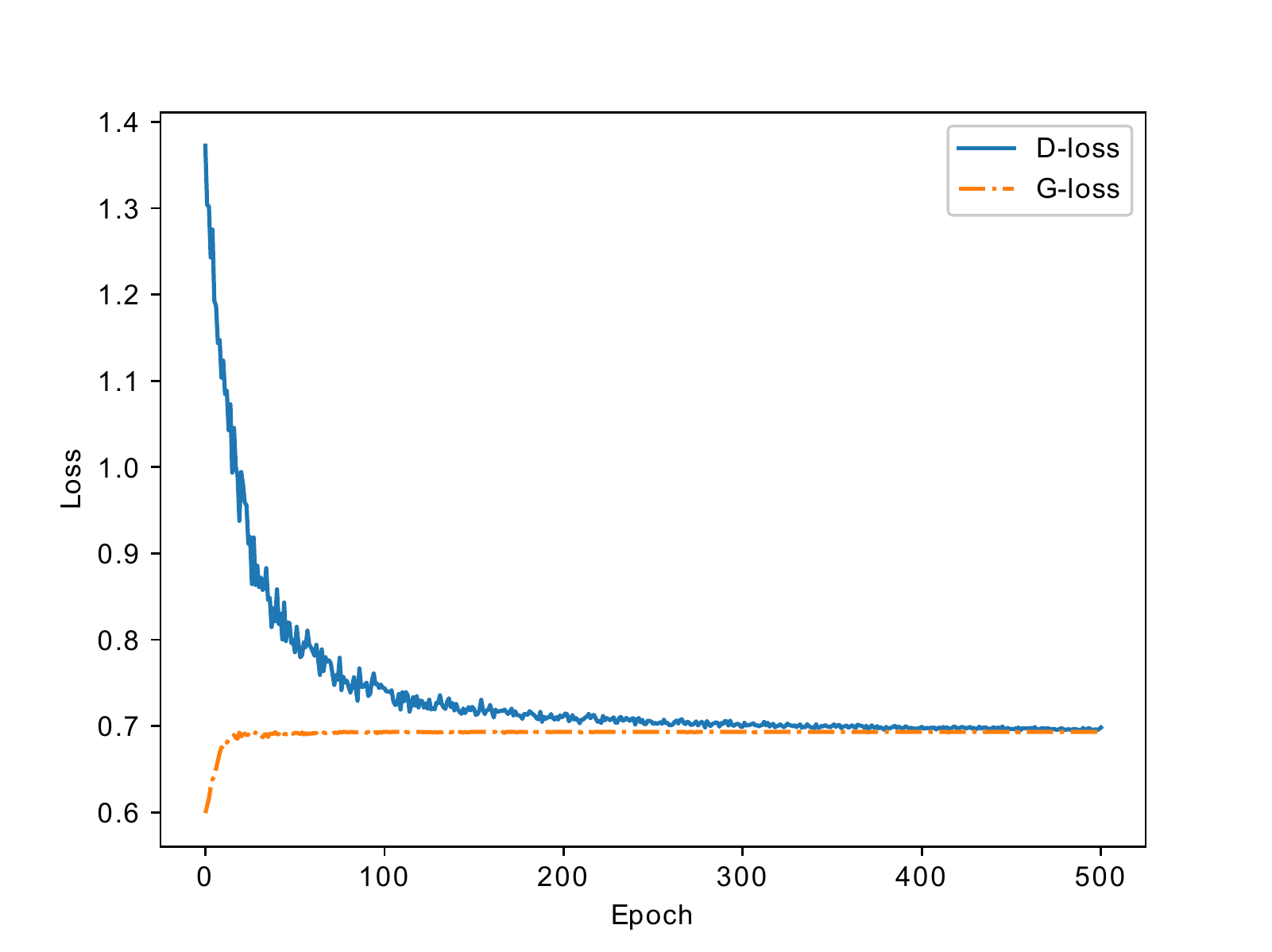}
	\caption{Loss function of $G$ and $D$ as a function of the training epochs.}\label{fig:dloss}
\end{figure}

In data augmentation, the objective of the adversary is to use $N_{r}$ real training data samples with labels, generate $N_{s}$ synthetic training data samples with labels by applying a conditional GAN, and augment the training data set to the total of $N_{aug} = N_{r}+N_{s}$ samples. The deep neural networks to build the conditional GAN are developed in TensorFlow. The generator and the discriminator are made up of FNNs with the following features.
\begin{itemize}
	\item The generator $G$ takes in a noise vector $\bm{z}$ that is randomly distributed with a zero mean and unit variance of $100$~dimensions, i.e., $\bm{z} \in \mathbb{R}^{100}$, and binary labels $\bm{y} \in \mathbb{R}^{2}$ as input, and generates synthetic data $G(\bm{z}|\bm{y})$ with the same dimensions of the original data. The first two layers of $G$ have 100 and 500 neurons per layer, respectively, and they are activated by rectifying linear unit (ReLU) that performs $\max(x,0)$ operation on input $x$. The output layer of $G$ has the same dimensions as the original data and uses the hyperbolic tangent (\emph{tanh}) activation function.
\item The discriminator $D$ takes in the original data or the synthetic data generated by $G$. The first two layers of $D$ have 500 neurons each, which are ReLU activated. The last layer of $D$ has one neuron that is sigmoid activated.
\end{itemize}
Both $G$ and $D$ are trained with the Adam optimizer \cite{Kingma14:Adam} using a learning rate of $10^{-5}$.
For the exploratory attack, the adversary starts with $N_r = $ 100 samples in training data to build the classifier $\hat T$. 500 samples are used in test data.
The hyperparatemers of the deep neural network used for inferring the target classifier are optimized as follows:
\begin{itemize}
\item The number of hidden layers is 3.
\item The number of neurons per layer is 50.
\item The loss function is cross entropy.
\item The activation function in hidden layers is sigmoid.
\item The activation function in output layer is softmax.
\item All weights and biases are initially scaled by 3.
\item Input values are unit normalized in the first training pass.
\item The minibatch size is 25.
\item The momentum coefficient to update the gradient is 0.1.
\item The number of epochs per time slot is 10.
\end{itemize}
The difference between the labels returned by $T$ and $\hat T$ is found as $d_1(\hat T, T) = 44.59\%$, $d_2(\hat T, T) = 45.64\%$, and $d(\hat T, T) = 45.00\%$. The adversary applies the GAN to generate additional synthetic training data samples based on 100 real training data samples. Then the adversary retrains $\hat T$ with updated training data. The difference between labels returned by $T$ and $\hat T$ is shown in Table~\ref{tab:gan}.
The GAN is trained with 500 epochs with 32 samples in each batch. In one epoch, two discriminator updates and one generator update are performed, as suggested in \cite{Goodfellow14:GAN}. Fig.~\ref{fig:dloss} shows the loss function of the generator and the discriminator during training.

When additional $N_s = $ 50 or 100 training data samples are generated by the GAN (i.e., when the total training data size is $N_{aug} = $ 150 or 200 samples), the performance of $\hat T$ improves significantly. However, adding more data (e.g., $N_s = $ 150, 200 or 300 synthetic data samples) generated by the GAN cannot further improve $\hat T$. While synthetic data samples generated by the GAN can provide more information to train a classifier, their labels may also incur some error (i.e., labels are different from those by $T$). Thus, adding too many additional synthetic data samples from the GAN hurts the performance of $\hat T$. Results in Table~\ref{tab:gan} show that training data augmentation with 100 real and 100 synthetic samples reduces $d(\hat T, T)$ up to $27.40\%$, which is only $1.6\%$ different from the case when all 3000 real samples are used for training in Section~\ref{sec:Datumbox}.

\section{Causative and Evasion Attacks}
\label{sec:ceattacks}

After an exploratory attack, causative and evasion attacks are launched by using the inferred classifier $\hat T$.

\subsection{Causative Attack}

In a causative attack, the adversary provides incorrect training data such that the classifier is retained with wrong data and the accuracy of the updated classifier $\tilde T$ drops.
In the extreme case, the adversary may change labels of all data samples.
However, to avoid being detected, the adversary may not prefer switching labels of too many data samples.
Thus, it is important to select the best set of data samples and switch their labels.
This selection requires the knowledge of $T$, but the adversary does not know $T$'s algorithm or training data.
Therefore, a successful causative attack can be performed only if the adversary has performed an exploratory attack and obtained the inferred classifier $\hat T$ that is similar to $T$.

The adversary applies deep learning to build $\hat T$, which also provides a likelihood score in $[0,1]$ on the classification of
each sample.
If this score is less than a threshold, this sample is identified as label $1$, otherwise it is identified as label $2$.
This score measures the confidence of classification.
That is, if a sample has a score close to $0$ or $1$, its label assignment has a high confidence.
On the other hand, if a sample has a score close to the threshold, its label assignment has a low confidence. With $\hat T$, an adversary can perform a causative attack by following three steps:
\begin{enumerate}
\item The adversary calls $\hat T$ with a number of samples to and receives their scores and labels.

\item Suppose the adversary can change the labels of for $p\%$  of samples.
The adversary selects samples with top $\frac{p}{2} \%$ scores and samples with bottom $\frac{p}{2} \%$ scores.

\item The adversary switches labels of selected samples and sends all labels of samples to retrain the classifier.
\end{enumerate}
We measure the impact of a causative attack by comparing the outputs of the original classifier $T$ and the updated classifier $\tilde T$ on samples with label $1$ or label $2$ (identified by $T$) and on all data samples.
Suppose there are $n_1$ samples with label $1$ and $n_2$ samples with label $2$.
Among $n_1$ (or $n_2$) samples, $m_1$ (or $m_2$) samples are classified differently by $\tilde T$.
Thus, we have three difference measures to represent the difference of $T$ and $\hat T$: $d_1(T, \tilde T) = \frac{m_1}{n_1}, d_2(T, \tilde T) = \frac{m_2}{n_2}, \mbox{ and } d(T, \tilde T) = \frac{m_1+m_2}{n_1+n_2} .$
We use 1000 samples and set $p=10$.
For the classifier $\hat T$ built in Section~\ref{sec:gan} with 400 training samples, we obtain $d_1(T, \tilde T) = 49.52\%$, $d_2(T, \tilde T) = 45.41\%$, and $d(T, \tilde T) = 48\%$.
This result shows that the causative attack reduces the updated classifier $\tilde T$'s performance significantly.

\subsection{Evasion Attack}

In an evasion attack, the adversary identifies a particular set of data samples that the classifier $T$ cannot reliably classify.
The adversary uses the classification scores by $\hat T$ to perform the evasion attack.
 If the objective of the attack is to maximize the error, the adversary should select samples with scores close to the threshold (i.e., with low confidence on classification). If the objective is to maximize the error of misclassifying a sample with label $1$ as label $2$ (or vice versa), the adversary should select samples classified as label $2$ (or $1$) and with score close to the threshold.
Although the evasion attack is performed with the inferred classifier, 
the evaluation of the attack's performance requires the ground truth on labels.
Since the adversary does not have access to the knowledge of ground truth, we cannot measure the performance of the evasion attack.

\section{Conclusion}
\label{sec:conclusion}
In this paper, we addressed adversarial deep learning with limitations on the training data. We considered the exploratory attack to infer an online classifier by training deep learning with a limited number of calls to a real online API.
However, with limited training data, we observed that there is a significant difference between the labels returned by the inferred classifier and the target classifier. We designed an approach using the GAN to reduce this difference and enhance the exploratory attack. 
Building upon the classifier inferred by the exploratory attack, we designed causative and evasion attacks to select training and test data samples, respectively, to poison the training process and fool the target classifier into making wrong decisions for these samples. Our results show that adversarial deep learning is feasible as a practical threat to online APIs even when only limited training data is used by the adversary.
\end{changemargin}


\begin{thebibliography}{99}


\bibitem{Papernot16}
N. Papernot, P. D. McDaniel, A. Sinha, and M. P. Wellman,
``Towards the science of security and privacy in machine learning,"
\emph{arXiv preprint arXiv:1611.03814}, 2016.

\bibitem{Goodfellow14}
I. J. Goodfellow, J. Shlens, and C. Szegedy,
``Explaining and harnessing adversarial examples,"
\emph{arXiv preprint arXiv:1412.6572},
2014.


\bibitem{Google:Vision}
Google Cloud Vision API,
available at https://cloud.google.com/vision.


\bibitem{Barreno06}

M. Barreno, B. Nelson, R. Sears, A. Joseph, and J. Tygar, ``Can machine learning be secure?" \emph{ACM Symposium on Information, Computer and Communications Security}, 2006.

\bibitem{Tramer16}
F. Tramer, F. Zhang, A. Juels, M. Reiter, and T. Ristenpart, ``Stealing machine learning models via prediction APIs," \emph{USENIX Security Symposium}, 2016.

\bibitem{Papernot17}
N. Papernot, P. McDaniel, I. Goodfellow, S. Jha, Z. Celik, and A. Swami,
``Practical black-box attacks against deep learning systems using adversarial examples,"
\emph{ACM Conference on Computer and Communications Security},
2017.

\bibitem{Shi17:DL}
Y.~Shi, Y. E.~Sagduyu, and A.~Grushin,
``How to steal a machine learning classifier with deep learning,"
\emph{IEEE Symposium on Technologies for Homeland Security (HST)},
2017.

\bibitem{Wu16}
X. Wu, M. Fredrikson, S. Jha, and J. F. Naughton,
``A methodology for formalizing model-inversion attacks," \emph{Computer Security Foundations},
2016.

\bibitem{bookchapter2018}
Y. Shi, Y. E. Sagduyu, K. Davaslioglu, and R. Levy, ``Vulnerability detection and analysis in adversarial deep learning," in \emph{Guide to Vulnerability Analysis for Computer Networks and Systems: An Artificial Intelligence Approach}, Springer, 2018.

\bibitem{Yi2018}
Y. Shi, Y. E Sagduyu, T. Erpek, K. Davaslioglu, Z. Lu, and J. Li,
``Adversarial deep learning for cognitive radio security: Jamming attack and defense strategies,''
\emph{IEEE International Conference on Communications (ICC) Workshop on Promises and Challenges of Machine Learning in Communication Networks}, 2018.

\bibitem{Pi16}
L. Pi, Z. Lu, Y. Sagduyu, and S. Chen,
``Defending active learning against adversarial inputs in automated document classification,"
\emph{IEEE Global Conference on Signal and Information Processing (GlobalSIP)},
2016.

\bibitem{Alfed16}
S. Alfeld, X. Zhu, and P. Barford,
``Data poisoning attacks against autoregressive models,"
 \emph{AAAI Conference on Artificial Intelligence},
2016.

\bibitem{Shi17:Milcom}
Y. Shi and Y. E Sagduyu,
``Evasion and causative attacks with adversarial deep learning,"
\emph{IEEE Military Communications Conference (MILCOM)},
2017.

\bibitem{Papernot16:evasion}
N. Papernot, P. McDaniel,  S. Jha, M. Fredrikson, Z. Celik, and A. Swami,
``The limitations of deep learning in adversarial settings,"
 \emph{IEEE European Symposium on Security and Privacy}, 2016.
2016.

\bibitem{Kurakin16}
A. Kurakin, I. Goodfellow, and S. Bengio,
``Adversarial examples in the physical world,"
\emph{arXiv preprint arXiv:1607.02533},
2016.

\bibitem{Moosavi-Dezfooli15:DeepFool}
S. M. Moosavi-Dezfooli, A. Fawzi, and P. Frossard,
``DeepFool: A simple and accurate method to fool deep neural networks,"
\emph{IEEE Conference on Computer Vision and Pattern Recognition},
2015.

\bibitem{Datumbox}
DatumBox Machine Learning API,
available at http://www.datumbox.\\
com/machine-learning-api/

\bibitem {Goodfellow14:GAN}
I. Goodfellow, J. Pouget-Abadie, M. Mirza, B. Xu, D. Warde-Farley, S. Ozair, A. Courville, and Y. Bengio,
``Generative adversarial nets,"
\emph{Advances in Neural Information Processing Systems},
2014.

\bibitem {DAGAN}
A. Antoniou, A. Storkey, and H. Edwards. ``Data augmentation generative adversarial networks," \emph{arXiv preprint arXiv:1711.04340}, 2017.

\bibitem{Kemal2018}
K. Davaslioglu and Y. E. Sagduyu,
``Generative adversarial learning for spectrum sensing,"
\emph{IEEE International Conference on Communications (ICC)}, 2018.

\bibitem{Tugba2018}
T. Erpek, Y. E. Sagduyu, and Y. Shi,
``Deep learning for launching and mitigating wireless jamming attacks,"
\emph{arXiv preprint arXiv:1807.02567}, 2018.

\bibitem{Yihst2018}
Y. Shi, Y. E. Sagduyu, K. Davaslioglu, and J. Li, ``Active Deep Learning Attacks under Strict Rate Limitations for Online API Calls," \emph{IEEE Symposium on Technologies for Homeland Security}, 2018

\bibitem{Arjovsky17}
M. Arjovsky and L. Bottou,
``Towards principled methods for training generative adversarial networks,"
\emph{International Conference on Machine Learning (ICLR)},
2017.

\bibitem{Mirza14:GAN}
M. Mirza and S. Osindero,
``Conditional generative adversarial nets,"
\emph{arXiv preprint arXiv:1411.1784},
2014.

\bibitem{Kingma14:Adam}
D. P. Kingma and J. Ba,
``Adam: A method for stochastic optimization,"
\emph{arXiv preprint arXiv:1412.6980},
2014.
\end{thebibliography}
\end{document}